\documentclass[lettersize,journal]{IEEEtran}

\usepackage[colorlinks,urlcolor=blue,linkcolor=blue,citecolor=blue]{hyperref}
\usepackage{amsmath,amsfonts}
\usepackage{amsthm}
\usepackage{array}
\usepackage{textcomp}
\usepackage{stfloats}
\usepackage{url}
\usepackage{verbatim}
\usepackage{graphicx}
\usepackage{cite}
\usepackage{booktabs}
\usepackage{multirow}
\usepackage{threeparttable}
\usepackage{dcolumn}
\usepackage{bm}
\usepackage{array}
\usepackage{enumerate}
\usepackage{siunitx}
\usepackage{subcaption}
\usepackage{caption}
\usepackage{float}
\usepackage{colortbl}
\usepackage{xcolor}
\usepackage{tabularx}
\usepackage{amssymb}
\usepackage[ruled,linesnumbered]{algorithm2e}

\begin{document}

\title{GaussFly: Contrastive Reinforcement Learning for Visuomotor Policies in 3D Gaussian Fields}

\author{Yuhang Zhang, Mingsheng Li, Yujing Shang, Zhuoyuan Yu, Chao Yan, Jiaping Xiao, and Mir Feroskhan, \IEEEmembership{Member, IEEE}
        
\thanks{Y. Zhang, M. Li, J. Xiao, and M. Feroskhan are with the School of Mechanical and Aerospace Engineering, Nanyang Technological University, Singapore 639798, Singapore (e-mail: yuhang004@e.ntu.edu.sg; m230168@e.ntu.edu.sg; jiaping001@e.ntu.edu.sg; mir.feroskhan@ntu.edu.sg). Y. Shang is with the School of Electrical and Electronic Engineering, Nanyang Technological University, Singapore 639798, Singapore (e-mail: yshang005@e.ntu.edu.sg). Z. Yu is with the College of Design and Engineering, National University of Singapore, Singapore 119077, Singapore (e-mail: yuzhuoyuan@u.nus.edu). C. Yan is with the College of Automation Engineering, Nanjing University of Aeronautics and Astronautics, Nanjing, 211106, China (e-mail: yanchao@nuaa.edu.cn). \textit{(Yuhang Zhang and Mingsheng Li contributed equally to this work.)} \textit{(Corresponding author: Mir Feroskhan)}

Project materials and supplementary information are available at: \url{https://zzzzzyh111.github.io/GaussFly_Web/}.}
}

\markboth{Journal of \LaTeX\ Class Files,~Vol.~14, No.~8, August~2021}%
{Shell \MakeLowercase{\textit{et al.}}: A Sample Article Using IEEEtran.cls for IEEE Journals}


\maketitle

\begin{abstract}
Learning visuomotor policies for Autonomous Aerial Vehicles (AAVs) relying solely on monocular vision is an attractive yet highly challenging paradigm. Existing end-to-end learning approaches directly map high-dimensional RGB observations to action commands, which frequently suffer from low sample efficiency and severe sim-to-real gaps due to the visual discrepancy between simulation and physical domains. To address these long-standing challenges, we propose GaussFly, a novel framework that explicitly decouples representation learning from policy optimization through a cohesive real-to-sim-to-real paradigm. First, to achieve a high-fidelity real-to-sim transition, we reconstruct training scenes using 3D Gaussian Splatting (3DGS) augmented with explicit geometric constraints. Second, to ensure robust sim-to-real transfer, we leverage these photorealistic simulated environments and employ contrastive representation learning to extract compact, noise-resilient latent features from the rendered RGB images. By utilizing this pre-trained encoder to provide low-dimensional feature inputs, the computational burden on the visuomotor policy is significantly reduced while its resistance against visual noise is inherently enhanced. Extensive experiments in simulated and real-world environments demonstrate that GaussFly achieves superior sample efficiency and asymptotic performance compared to baselines. Crucially, it enables robust and zero-shot policy transfer to unseen real-world environments with complex textures, effectively bridging the sim-to-real gap.
\end{abstract}


\section{Introduction}
The widespread deployment of Autonomous Aerial Vehicles (AAVs) in complex and unstructured environments relies on the capability to achieve robust navigation under strict hardware constraints \cite{loquercio2021learning}. Among various sensing modalities, monocular vision stands out as the perceptual channel most closely aligned with human perception. As a result, monocular visual navigation has long been regarded as an attractive yet challenging paradigm for autonomous flight. Despite its inherent difficulty, the proficiency of expert human pilots provides compelling evidence of its feasibility. Even in the presence of transmission latency and signal noise, human operators can execute complex maneuvers relying exclusively on first-person view (FPV) feedback. This capability is fundamentally driven by the cognitive ability to intuitively perceive geometric structure and spatial context directly from visual observations, rather than merely processing pixel intensities. Therefore, in this work, we ask: How can AAVs achieve human-like robust flight relying solely on monocular vision?

Admittedly, active ranging sensors, such as LiDAR and RGB-D cameras, simplify spatial perception by providing direct depth measurements. However, they generally incur higher weight and power costs compared to monocular solutions, restricting their efficiency on lightweight AAVs. Monocular cameras offer a more compact alternative but fundamentally shift the burden from hardware to perception. Conventional approaches typically rely on explicit geometric reconstruction pipelines, such as Simultaneous Localization and Mapping (SLAM) \cite{matsuki2024gaussian} and Structure from Motion (SfM) \cite{wang2024tc}. These methods operate by extracting and matching geometric features to estimate positions and reconstruct the environment. However, such processes often demand substantial computational resources that strain the limited onboard capacity. As a result, classical geometric methods are insufficient to fully exploit the rich perceptual potential of monocular vision.

To overcome the aforementioned limitations, data-driven learning has emerged as a competitive alternative \cite{xiao2025vision}. Broadly, these approaches can be categorized into imitation learning (IL) from real-world datasets \cite{gandhi2017learning, loquercio2018dronet, kouris2018learning} and reinforcement learning (RL) in simulation \cite{kaufmann2023champion, zhang2024npe, zhang2025learning1}. While learning from real-world data ensures visual fidelity, collecting large-scale, diverse expert trajectories is prohibitively labor-intensive and risky for aerial platforms. Consequently, researchers increasingly resort to training visuomotor policies in virtual environments. However, learning directly from raw monocular images in simulation introduces two fundamental challenges. First, unlike depth maps or point clouds that offer direct geometric measurements, RGB images inherently lack explicit depth information. This necessitates the implicit extraction of spatial geometry from image input, which is a process that is significantly more ambiguous than processing range data. Second, the substantial visual discrepancy between synthetic and physical environments results in a severe sim-to-real gap. Standard simulators struggle to perfectly replicate the complex lighting dynamics, texture details, and sensor noise of the physical world. This visual distribution shift often causes policies that perform flawlessly in simulation to degrade or fail completely when deployed in real-world scenarios.

To bridge the sim-to-real gap, existing methods mainly fall into two categories: domain randomization \cite{tobin2018domain, horvath2022object} and the use of explicit intermediate representations \cite{wang2025navbest, hu2025seeing}. Domain randomization aims to cover real-world distributions by randomizing visual and physical parameters in simulation. While effective, this approach often requires manual tuning of randomization ranges and tends to yield conservative policies that sacrifice flight agility to adapt to unrealistic synthetic variations. Alternatively, other approaches decouple perception from policy learning by predicting geometric modalities, such as depth map \cite{wang2025navbest} or optical flow \cite{hu2025seeing}. However, relying on dense depth estimation introduces high computational latency and is sensitive to accumulated estimation errors. Similarly, optical flow tends to lose rich semantic context and suffer from ambiguity when distinguishing ego-motion from environmental changes.

In this paper, we introduce GaussFly, a novel visuomotor policy learning framework designed to address the aforementioned challenges through a cohesive real-to-sim-to-real paradigm. First, to achieve a high-fidelity real-to-sim transition and bridge the visual discrepancy between synthetic and physical domains, we establish a training environment by reconstructing real-world scenes using 3D Gaussian Splatting (3DGS) \cite{kerbl20233d}. We enforce explicit planar constraints and normal consistency during optimization, ensuring that the reconstructed environments are not only photorealistic but also geometrically accurate. Second, to ensure robust sim-to-real transfer, we explicitly decouple representation learning from policy learning. Specifically, based on the reconstructed scenes, we collect a dataset containing diverse views and introduce a contrastive representation learning approach to train a visual encoder. By aligning latent representations of spatially corresponding images across drastically different viewpoints and illumination conditions, this approach allows the encoder to isolate task-relevant visual features from background noise, reducing the reliance on reward signals \cite{stooke2021decoupling}. Finally, this encoder is frozen and integrated into an RL framework, where it maps high-dimensional visual observations into compact state representations for the policy to generate action commands. GaussFly significantly improves sample efficiency and asymptotic performance. Furthermore, it achieves successful zero-shot transfer back to the physical world, demonstrating navigation capabilities across real-world scenarios with different obstacle configurations and complex textural variations.

The contributions of this work are outlined as follows:
\begin{enumerate}
    \item We propose GaussFly, a novel visuomotor policy learning framework designed for AAVs, which effectively addresses the limitations of monocular perception and the severe sim-to-real gap. By bridging high-fidelity simulation with geometric representation learning, GaussFly enables robust autonomous flight only using RGB inputs.
    
    \item We establish a geometrically consistent simulation environment using 3DGS with explicit planar constraints and normal consistency regularization. Our approach can achieve real-time rendering of visual inputs at 30\,Hz.
    
    \item We introduce a decoupled contrastive representation learning approach that aligns latent representations of spatially corresponding images across diverse viewpoints and illumination conditions. It forces the visual encoder to distill task-relevant visual features while filtering out background noise, which improves the sample efficiency.
    
    \item We conduct comprehensive experiments in simulation and real-world environments. Results demonstrate that GaussFly achieves superior asymptotic performance compared to all baselines and enables zero-shot policy transfer to various real-world environments.
\end{enumerate}

\section{Related Work}
This section divides related work into two categories: visuomotor policy learning and contrastive representation learning.

\subsection{Visuomotor Policy Learning}
Visuomotor policy learning for AAVs is mainly categorized into IL and RL. Early approaches \cite{gandhi2017learning, kouris2018learning, loquercio2018dronet} largely adopted the IL paradigm, utilizing supervised learning to regress control actions from large-scale expert datasets. These methods demonstrated feasibility in structured environments. However, their efficiency is limited by the diversity and volume of the training data, creating a significant scalability bottleneck for complex and unstructured tasks. To mitigate the heavy reliance on expert demonstrations, RL has gained traction \cite{kaufmann2023champion, zhang2024npe, zhang2025learning1, yan2023collision, xiao2023collaborative}. A prominent example is the work of Kaufmann et al. \cite{kaufmann2023champion}, who developed a champion-level racing policy using onboard cameras. Their system maps visual inputs directly to control commands to achieve superhuman agility. More recently, Zhang et al. \cite{zhang2025learning1} introduced a differentiable simulator for AAVs. By integrating depth modalities with first-order gradients, they achieved agile and safe flight in cluttered environments. Despite these successes, the sim-to-real gap remains a major challenge for monocular RL. Due to the difficulty of transferring policies from simulation to the real world, the aforementioned works often resort to depth maps or explicit intermediate representations (e.g., gate edges), rather than raw RGB inputs, leaving the potential of monocular visuomotor policy learning underexplored.

The emergence of 3DGS has recently revolutionized scene representation, prompting investigations into its utility for visuomotor policy learning. However, current applications remain in early stages. Some existing research \cite{quach2024gaussian, tagliabue2024tube, low2025sous} leverages 3DGS primarily as a high-fidelity data generator to augment IL datasets, failing to escape the fundamental limitations of supervised learning. While \cite{chen2025grad} has integrated 3DGS with RL, it is confined to simplified tasks such as point-to-point traversal and relies heavily on dense reward signals derived from explicit trajectory waypoints. Although a recent framework \cite{huang2025flying} explored end-to-end navigation using 3DGS-rendered images, it primarily relies on domain randomization within limited scenarios to facilitate transfer, rather than focusing on improving feature extraction from RGB images.

\subsection{Contrastive Representation Learning}
Unlike approaches that generate explicit intermediate representations, such as depth maps \cite{wang2025navbest} or point clouds \cite{hu2025seeing}, contrastive representation learning \cite{fu2023learning, choi2024efficient, xing2024contrastive, zhang2025learning, zhang2025oracle} adopts an implicit paradigm. It aims to extract compact, low-dimensional representations directly from raw sensor inputs. Since this paradigm significantly reduces computational overhead compared to pixel-wise estimation, it has been widely adopted in visuomotor policy learning. For instance, Fu et al. \cite{fu2023learning} combined object detection algorithms with representation learning to extract task-relevant features from RGB images for drone racing. These features were subsequently utilized to distill a student policy from a privileged teacher, enhancing performance in complex environments. To address the sensitivity of RGB images to texture and illumination changes, Zhang et al. \cite{zhang2025learning} proposed a cross-modal contrastive framework. By aligning RGB embeddings with depth representations in a shared latent space, their method extracts depth-consistent latent representations that encode structural cues of the environment. These representations remain resilient to visual noise, thereby improving the robustness of the visuomotor policy. Building on this, Zhang et al. \cite{zhang2025oracle} incorporated masking techniques and temporal modeling to further refine representation learning. By considering the time horizon of the visuomotor policy and utilizing Transformer architectures to reconstruct randomly masked RGB inputs in the latent space, this approach effectively extracts task-relevant yet scene-agnostic features.

Despite these advancements, the training of feature extractors largely relies on data collected in high-fidelity simulators, which only partially mitigates the sim-to-real gap. While fine-tuning on real-world datasets is a feasible solution, it still incurs high data collection costs. Consequently, research exploring the integration of contrastive representation learning with photorealistic and geometrically consistent reconstructions of real-world scenes remains limited.

\section{Preliminaries and Problem Formulation}
This section presents the preliminaries of  GaussFly, including the problem formulation and the fundamentals of 3DGS.

\subsection{Problem Formulation}
Considering the restricted FPV and the absence of explicit state estimation, we model the visuomotor policy learning for AAVs as a Partially Observable Markov Decision Process (POMDP). This process is defined by the tuple $(\mathcal{O}, \mathcal{A}, \mathcal{P}, \mathcal{R}, \gamma)$. Here, $\mathcal{O}$ represents the observation space, $\mathcal{A}$ is the agent's action space, $\mathcal{P}$ denotes the transition dynamics, $\mathcal{R}: \mathcal{O} \times \mathcal{A} \rightarrow \mathbb{R}$ is the reward function, and $\gamma \in [0,1)$ is the discount factor. Unlike standard formulations where $\mathcal{O}$ consists of high-dimensional image inputs \cite{chen2025grad, huang2025flying}, we explicitly decouple perception from policy learning. Let $\mathbf{I}_t \in \mathbb{R}^{H \times W \times 3}$ denote the raw monocular RGB input at timestep $t$. We introduce a visual encoder $E_\phi$ and a subsequent non-linear projection head $P_\psi$, to map the raw input into a compact latent observation space $\mathcal{O} \subset \mathbb{R}^d$. Consequently, the observation at each step is defined as the extracted feature vector $\mathbf{o}_t = P_\psi(E_\phi(\mathbf{I}_t))$. Due to partial observability, a single observation $\mathbf{o}_t$ is insufficient to infer the full system dynamics. The agent must rely on a history of past interactions. We define the history buffer at timestep $t$ as $\mathbf{h}_t = (\mathbf{o}_{t-k}, \dots, \mathbf{o}_t)$, where $k$ is the temporal window size. The objective is to learn a visuomotor policy $\pi_\theta: \mathcal{H} \rightarrow \mathcal{A}$, parameterized by $\theta$, that maps the history $\mathbf{h}_t \in \mathcal{H}$ to an optimal action $\mathbf{a}_t$. The optimization objective is to maximize the expected cumulative discounted return: $J(\pi_\theta) = \mathbb{E}_{\pi_\theta} \left[ \sum_{t=0}^{T} \gamma^t \mathcal{R}(\mathbf{o}_t, \mathbf{a}_t) \right]$.

\subsection{3DGS Formulation}
To minimize the visual and geometric discrepancy between simulation and real-world environments, we reconstruct photorealistic training scenes using 3DGS. In this representation, a scene is modeled as a set of 3D Gaussian primitives, each parameterized by a mean position $\boldsymbol{\mu}_i \in \mathbb{R}^3$, a covariance matrix $\boldsymbol{\Sigma}_i \in \mathbb{R}^{3 \times 3}$, view-dependent radiance coefficients, and an opacity value. A Gaussian primitive defines a continuous volumetric density function:
\begin{equation}
G_i(\mathbf{x}) = \exp\!\left(-\frac{1}{2}(\mathbf{x}-\boldsymbol{\mu}_i)^\top \boldsymbol{\Sigma}_i^{-1}(\mathbf{x}-\boldsymbol{\mu}_i)\right).
\end{equation}

Following the standard rendering pipeline, the final pixel color is obtained via front-to-back volumetric compositing of all Gaussians that are projected to the pixel. The rendered color $C$ is computed as:
\begin{equation}
C = \sum_{i \in \mathcal{N}} c_i \alpha_i \prod_{j=1}^{i-1} (1 - \alpha_j),
\end{equation}
where $\mathcal{N}$ denotes the ordered set of contributing Gaussians, $c_i$ is the view-dependent radiance of Gaussian $i$, and $\alpha_i \in [0,1]$ is its effective opacity.

\section{Methodology}
This section presents the proposed framework, including an overview, scene reconstruction, contrastive representation learning, and policy learning.

\subsection{Overview}
The overall architecture of the proposed GaussFly framework is illustrated in Fig. \ref{fig:framework}. To effectively bridge the sim-to-real gap, it explicitly decouples representation learning from policy optimization through a cohesive real-to-sim-to-real pipeline. First, we employ 3DGS and a frozen SAM2 model to reconstruct and independently compose background environments and foreground assets, generating diverse, high-fidelity simulation environments. Second, we introduce contrastive representation learning to train a visual encoder. By minimizing an InfoNCE loss ($\mathcal{L}_{\text{InfoNCE}}$) to align latent representations of differently augmented on-the-fly renderings, the encoder learns to extract noise-resilient visual features. Finally, this optimized encoder is frozen to provide compact state observations for policy learning, which can be directly deployed for sim-to-real transfer.

\subsection{Scene Reconstruction}

While the original 3DGS formulation prioritizes visual fidelity, accurate geometric structure is critical for visuomotor learning and collision-aware navigation. To this end, we adopt a planar-constrained 3DGS variant that enforces geometric consistency during reconstruction \cite{chen2024pgsr}. Specifically, each Gaussian is encouraged to exhibit a planar structure by aligning its smallest covariance axis with the local surface normal. Let $\boldsymbol{\Sigma}_i = \mathbf{R}_i \mathbf{S}_i \mathbf{S}_i^\top \mathbf{R}_i^\top$ denote the eigendecomposition, where $\mathbf{R}_i$ is the rotation matrix and $\mathbf{S}_i=\mathrm{diag}(s_{i,x}, s_{i,y}, s_{i,z})$ represents the scaling factors. Without loss of generality, we assume $s_{i,z}$ corresponds to the smallest scale axis. We enforce the planarity by minimizing $s_{i,z}$ and define the normal direction as $\mathbf{n}_i = \mathbf{r}_{i,3}$, which corresponds to the third column of $\mathbf{R}_i$ aligned with the shortest axis. Based on this planar assumption, depth values are computed via ray–plane intersection rather than Gaussian center projection, yielding geometrically consistent depth maps. To enforce this structural fidelity, we explicitly minimize a geometric regularization term $\mathcal{L}_{\text{geo}}$ during optimization:
\begin{equation}
\label{eq:geo}
\mathcal{L}_{\text{geo}} = \lambda_{s} \sum_{i} s_{i,z} + \lambda_{n} \sum_{\mathbf{u}} \| \mathbf{n}_{\text{render}}(\mathbf{u}) - \mathbf{n}_{\text{depth}}(\mathbf{u}) \|_1,
\end{equation}
where $\lambda_s$ and $\lambda_n$ are weighting coefficients. The first term flattens the Gaussians along the surface normal, while the second term enforces consistency between the appearance-driven surface normals $\mathbf{n}_{\text{render}}(\mathbf{u})$, and the geometry-driven pseudo-normals $\mathbf{n}_{\text{depth}}(\mathbf{u})$ computed from ray-plane depth.

\begin{figure*}[!t]
   \centering
   \includegraphics[width=1.0\textwidth]{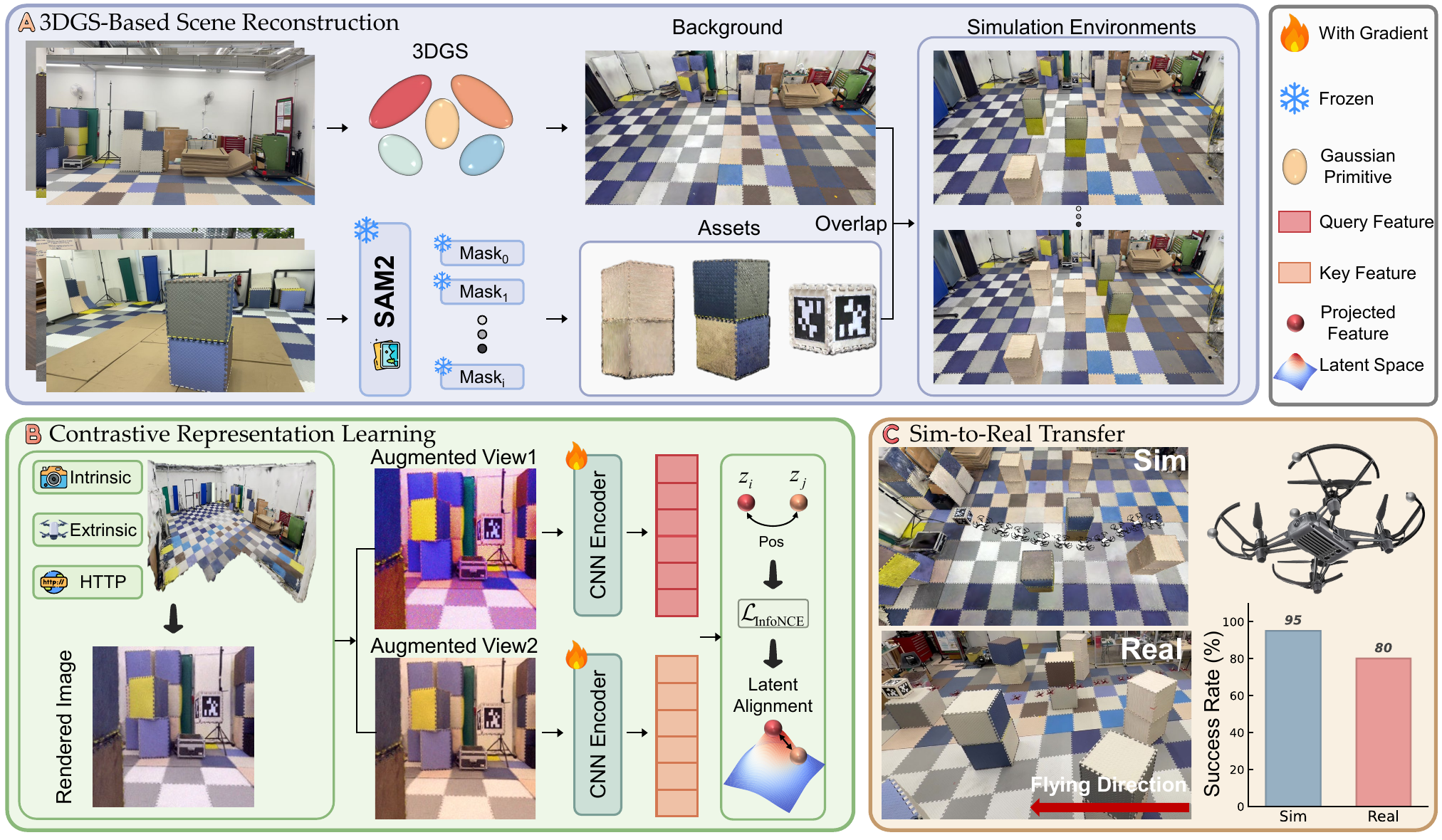} 
   \caption{The framework of GaussFly. (A) 3DGS-Based Scene Reconstruction. Background environments and foreground assets are reconstructed independently. By utilizing a frozen SAM2 model \cite{ravi2025sam} for asset masking, the framework enables flexible object composition and extensive domain randomization to generate diverse simulation environments. (B) Contrastive Representation Learning. Visual observations are rendered on-the-fly based on camera intrinsics and extrinsics. Two differently augmented views of the same rendered image are processed by CNN encoders, and an InfoNCE loss ($\mathcal{L}_{\text{InfoNCE}}$) is applied to align their latent representations, thereby extracting noise-resilient visual features. The optimized CNN encoder is subsequently frozen for policy learning. (C) Sim-to-Real Transfer. Benefiting from the high-fidelity 3DGS environments and robust latent representations, the trained visuomotor policy achieves seamless zero-shot transfer, demonstrating high navigation success rates across both simulation and real-world environments.}
   \label{fig:framework}
\end{figure*}

We reconstruct two distinct real-world scenes using video recordings. As shown in Fig. \hyperref[fig:framework]{\ref*{fig:framework}A}, to enable extensive domain randomization and flexible layout configurations, the background environments and foreground assets (i.e., obstacles and target goals) are reconstructed independently. Specifically, we employ the SAM2 model \cite{ravi2025sam} to accurately mask and extract these individual assets from the raw footage. The pipeline begins with COLMAP SfM \cite{schonberger2016structure} to initialize the point cloud and estimate camera poses, followed by 3DGS optimization supervised by Eq.~\eqref{eq:geo}. The decoupled optimized Gaussians are then scaled, composed, and aligned within the simulation coordinate system in Isaac Sim. Finally, we extract a continuous surface mesh from the composed scenes via Truncated Signed Distance Function (TSDF) fusion \cite{zhang2024neural} for precise collision checking. During training, RGB observations are rendered on-the-fly from the agent’s current position and orientation using the optimized 3DGS models at 30\,Hz.

\subsection{Contrastive Representation Learning}

To extract robust and compact state representations from raw RGB inputs without relying on explicit state estimation, and to further enhance scene understanding within the 3DGS-based environment, we introduce contrastive representation learning. GaussFly operates directly on the visual domain by maximizing the agreement between differently augmented views of the same rendered RGB input.

As shown in Fig. \hyperref[fig:framework]{\ref*{fig:framework}B}, given a minibatch $\mathcal{B}$ of $N$ raw images rendered from the 3DGS environment, we apply a stochastic data augmentation module $\mathcal{T}$ to generate two different views for each image $\mathbf{I}_t$. Let $\tilde{\mathbf{I}}_i$ and $\tilde{\mathbf{I}}_j$ denote two different augmented versions of the same source image, forming a positive pair. The remaining augmented images within the minibatch are treated as negative samples. The visual encoder $E_\phi$ first maps these augmented inputs into a high-dimensional representation space, yielding intermediate features $f_i = E_\phi(\tilde{\mathbf{I}}_i)$ and $f_j = E_\phi(\tilde{\mathbf{I}}_j)$. To further enhance the expressiveness of the contrastive alignment, these representations are subsequently processed by a non-linear projection head $P_\psi(\cdot)$, yielding the final projected query feature $z_i = P_\psi(\mathbf{h}_i)$ and key feature $z_j = P_\psi(\mathbf{h}_j)$ in the latent space. The objective is to pull positive pairs closer in the latent space while pushing negative pairs apart. From an information-theoretic viewpoint, the above objective aligns with the Joint Information Bottleneck (JIB) principle \cite{federici2020learning}, which aims to extract a latent representation that preserves shared, task-relevant information across augmented views while suppressing augmentation-induced noise. Since directly optimizing the JIB objective is intractable for high-dimensional visual inputs, we introduce the Information Noise-Contrastive Estimation (InfoNCE) loss \cite{chen2020simple} as a practical lower-bound estimator of mutual information, enabling efficient contrastive optimization. For a given positive pair of representations $(\mathbf{z}_i, \mathbf{z}_j)$, the loss is defined as:
\begin{equation}
\label{eq:infonce}
\mathcal{L}_{\text{InfoNCE}} = - \mathbb{E}_{\mathcal{B}} \left[ \log \frac{\exp(\text{sim}(\mathbf{z}_i, \mathbf{z}_j) / \tau)}{\sum_{k=1}^{N} \exp(\text{sim}(\mathbf{z}_i, \mathbf{z}_k) / \tau)} \right],
\end{equation}
where $\tau$ is a temperature parameter that controls the sharpness of the distribution. This formulation forces the encoder to capture features that are invariant to visual perturbations. $\text{sim}(\cdot, \cdot)$ is computed using cosine similarity:
\begin{equation}
\text{sim}(\mathbf{u}, \mathbf{v}) = \frac{\mathbf{u} \cdot \mathbf{v}}{\|\mathbf{u}\| \|\mathbf{v}\|}.
\end{equation}

The feature encoder $E_\phi$ is instantiated as a ResNet-50 backbone \cite{he2016deep}, while the non-linear projection head $P_\psi$ is implemented as a two-layer multi-layer perceptron (MLP) with a hidden ReLU activation. During the pre-training phase, the data augmentation pipeline includes random cropping, resizing, horizontal flipping, and Gaussian blurring. The model is optimized using stochastic gradient descent to minimize Eq.~\eqref{eq:infonce}. Once the contrastive pre-training is complete, the parameters $\phi$ of the encoder and $\psi$ of the projection head are frozen and employed as the static feature extractor for the subsequent visuomotor policy learning.

\subsection{Policy Learning}

The visuomotor policy maps visual observations to continuous action commands for autonomous flight. The pre-trained visual encoder $E_\phi$ and the non-linear projection head $P_\psi$ are kept frozen, jointly serving as a fixed feature extraction pipeline. This composite module transforms monocular RGB observations into compact latent representations that are subsequently fed into the visuomotor policy.

We train the policy using the Proximal Policy Optimization (PPO) algorithm \cite{schulman2017proximal}, which is optimized by minimizing the following objective:

\begin{equation}
\label{eq:ppo_actor_entropy}
\begin{aligned}
\mathcal{L}_{\text{PPO}}
=
& -\mathbb{E}_t \left[ 
\min \left( r_t \hat{A}_t,\ \text{clip}(r_t, 1 - \epsilon, 1 + \epsilon) \hat{A}_t \right) 
\right] \\
& + \mathbb{E}_t \left[ \left( V(o_t) - \hat{V}_t \right)^2 \right],
\end{aligned}
\end{equation}
where $r_t$ is the probability ratio between new and old policies, $\hat{A}_t$ denotes the advantage estimate, and $\epsilon$ is the clipping range. In addition, $V(o_t)$ denotes the predicted state value based on the current latent observation, while $\hat{V}_t$ corresponds to the target return derived from empirical Monte Carlo sampling.

At each timestep \( t \), the policy receives a composite observation consisting of three temporally consecutive visual feature embeddings \( \{\mathbf{z}_{t-2}, \mathbf{z}_{t-1}, \mathbf{z}_{t}\} \), the agent’s proprioceptive state \( \mathbf{s}_t \) (including orientation, linear velocity, and angular velocity), and the relative position of the agent with respect to the target goal. Both the actor and critic networks share an identical architecture. Prior to fusion, visual features and proprioceptive inputs are independently normalized. The stacked visual features are concatenated with the normalized proprioceptive state and goal representation to form a joint state vector. This vector is subsequently processed by an MLP consisting of two hidden layers with 256 units, where ReLU activations are applied. The actor network outputs continuous action commands corresponding to the agent’s linear velocity \( \mathbf{v}_t \) and angular velocity \( \boldsymbol{\omega}_t \), which are clipped to predefined limits \( \mathbf{v}_{\max} \) and \( \boldsymbol{\omega}_{\max} \) before execution.

An episode terminates when one of the following conditions is met: (i) the agent reaches the target region, (ii) a collision with an obstacle occurs, or (iii) the maximum episode length $T_{\text{lim}}$ is exceeded. The reward function is designed to encourage efficient and collision-free navigation. Specifically, a terminal reward $r_{\text{goal}}$ is granted upon reaching the target, while a penalty $r_{\text{col}}$ is applied in the event of a collision. To promote time-efficient behavior, a small per-step penalty $r_{\text{step}} = -1 / T_{\text{lim}}$ is imposed. In addition, a shaping reward proportional to forward progress is introduced:
\begin{equation}
r_{\text{prog}} = \kappa \, (d_{t-1} - d_t),
\end{equation}
where $d_t$ denotes the Euclidean distance between the agent and the target at timestep $t$, and $\kappa$ is a scaling coefficient. The final reward is computed as the sum of these components.

\section{Experimental Setup}
This section presents the experimental setup used to evaluate GaussFly. We describe the simulation environments, evaluation metrics, baseline methods, and implementation details.

\begin{figure}[t!]
    \centering
    \begin{subfigure}[t]{0.48\columnwidth}
        \centering
        \includegraphics[width=\linewidth]{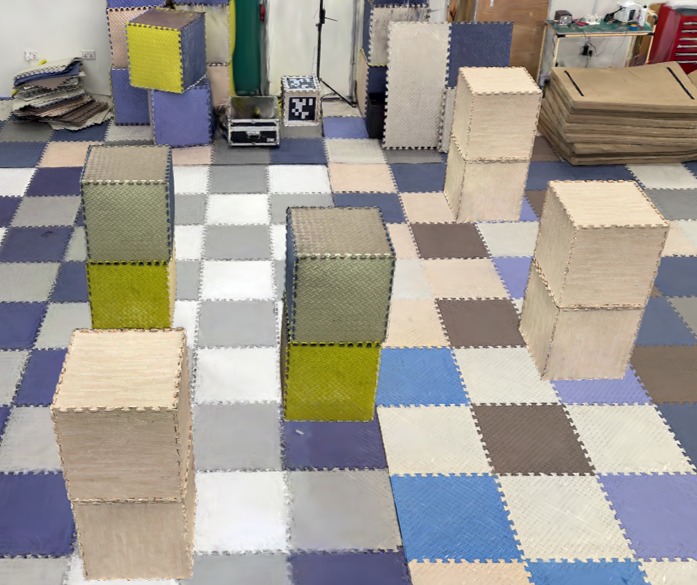}
        \caption{Scene A}
        \label{fig:scene1}
    \end{subfigure}
    \hfill
    \begin{subfigure}[t]{0.48\columnwidth}
        \centering
        \includegraphics[width=\linewidth]{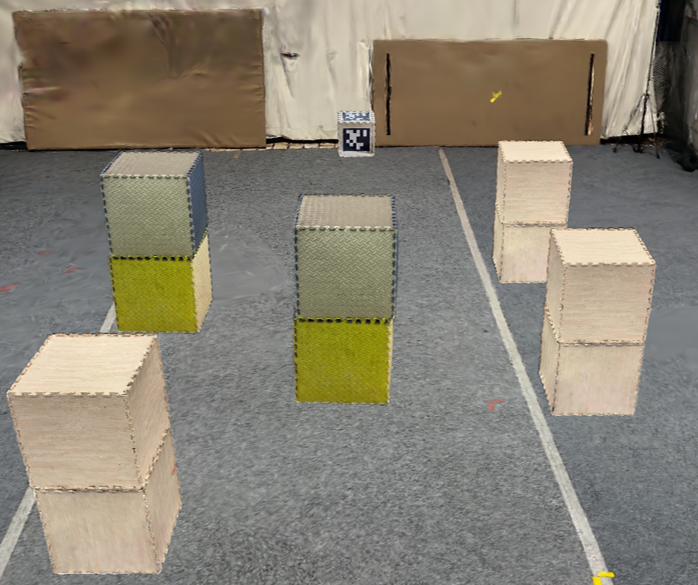}
        \caption{Scene B}
        \label{fig:scene2}
    \end{subfigure}
    \caption{Illustration of the simulation environments.}
    \label{fig:scene}
\end{figure}

\subsection{Simulation Setup}
We build the simulation environment in Isaac Sim, as shown in Fig. \ref{fig:scene}. Real-world scenes and objects are captured using an iPhone~17~Pro~Max by recording multi-view videos. Specifically, we reconstruct two distinct indoor scenes: Scene A, characterized by a cluttered background and a colorful floor, and Scene B, which presents a clean background with a uniform, solid-color floor. For each scene, we record approximately 4--5 minutes of video footage, while individual objects are captured with roughly 1 minute of close-range scanning. These videos are then used to reconstruct high-fidelity 3DGS representations. After reconstruction, the resulting meshes are imported into Isaac Sim for collision detection. Crucially, the reconstructed meshes are used for physics interactions, while visual observations are rendered on-the-fly using the optimized 3DGS models at 30\,Hz. In addition, domain randomization is applied to the rendered images, including variations in photometric appearance, illumination intensity, and color statistics. 

For contrastive representation learning, we collect $10,000$ RGB images rendered from the reconstructed environments at a resolution of $64 \times 64$. These images are used to pre-train the visual encoder $E_\phi$ and the projection head $P_\psi$. Specifically, the two-layer MLP of $P_\psi$ is configured with a hidden dimension of 256 and an output feature dimensionality of 128, using a temperature parameter $\tau = 0.07$. 

The visuomotor policy is trained in simulation and evaluated without fine-tuning across different environments. At each episode reset, obstacle layouts and target goal positions are randomized to encourage robust policy learning. For training, the batch size is set to 1024, the rollout horizon is 10,240, the learning rate is $3 \times 10^{-4}$, and the clipping parameter $\epsilon$ is set to $0.2$. The velocity limits are set to $\mathbf{v}_{\max}=1.5$\,m/s and $\boldsymbol{\omega}_{\max}=1.5$\,rad/s. For the reward setting, we set $r_{\text{goal}}=10$, $r_{\text{col}}=-1$, $T_{\text{lim}}=5000$, and $\kappa=0.1$. All simulation training is conducted for approximately eight hours on an NVIDIA RTX 6000 Ada GPU.

\subsection{Metrics}

We assess navigation performance using a set of standard metrics commonly adopted in visuomotor learning benchmarks~\cite{zhang2025grounded}. Specifically, we report the following:

\begin{itemize}
    
    \item \textit{Oracle Success (OS)}: the percentage of episodes in which the agent's trajectory intersects a success region within a radius of $\varepsilon = 0.5\,\mathrm{m}$ around the goal at any timestep.
    
    \item \textit{Success Rate (SR)}: the percentage of episodes in which the agent successfully reaches the goal.

    \item \textit{Collision Rate (CR)}: the percentage of episodes in which the agent collides with obstacles.

    \item \textit{Navigation Error (NE)}: the mean Euclidean distance $(\mathrm{m})$ between the agent's final position and the target location.

    \item \textit{Time to Success (TTS)}: the average number of steps required to reach the goal over successful episodes.
    
    \item \textit{Success weighted by Path Length (SPL)}: a composite metric that jointly evaluates navigation success and efficiency, defined as:
    \begin{equation}
    \mathrm{SPL} = \frac{1}{M} \sum_{i=1}^{M} \frac{\mathbb{I}_i \, \ell_i}{\max(d_i, \ell_i)},
    \end{equation}
    where $M$ is the total number of episodes, $\mathbb{I}_i$ indicates whether the $i$-th episode is successful, $\ell_i$ denotes the shortest-path distance to the goal, and $d_i$ is the length of the executed trajectory.

\end{itemize}

All metrics are reported in simulation under a strict success criterion, where an episode is considered successful only if the agent precisely reaches the goal location. In real-world experiments, we report only the success rate, and an episode is deemed successful if the agent arrives within a $0.5\,\mathrm{m}$ radius of the target to account for safety considerations.

\subsection{Baselines}
To comprehensively validate the performance of GaussFly, we compare it against several representative baselines spanning RL, IL, and traditional planning-based approaches.

\begin{itemize}
    \item D3QN \cite{zhang2023partially}: A value-based RL baseline that directly maps raw RGB observations to discrete control commands using a conventional convolutional neural network.

    \item PPO \cite{schulman2017proximal}: A policy-based RL baseline that directly maps raw RGB observations to continuous control commands using a conventional convolutional neural network.
    
    \item NPE \cite{zhang2024npe}: A hybrid RL framework that incorporates non-expert demonstrations to facilitate sim-to-real transfer in visuomotor policy learning, improving exploration efficiency while maintaining strong asymptotic performance.

    \item DAgger \cite{kelly2019hg}: An iterative IL algorithm designed to mitigate covariate shift by aggregating expert corrections over the learner’s induced state distribution.

    \item Hybrid APF \cite{pan2021improved}: A classical planning-based method that combines Artificial Potential Fields (APF) with A* search to alleviate local minimum issues.
\end{itemize}

\begin{figure}[t!]
    \centering
    \includegraphics[width=1.0\linewidth]{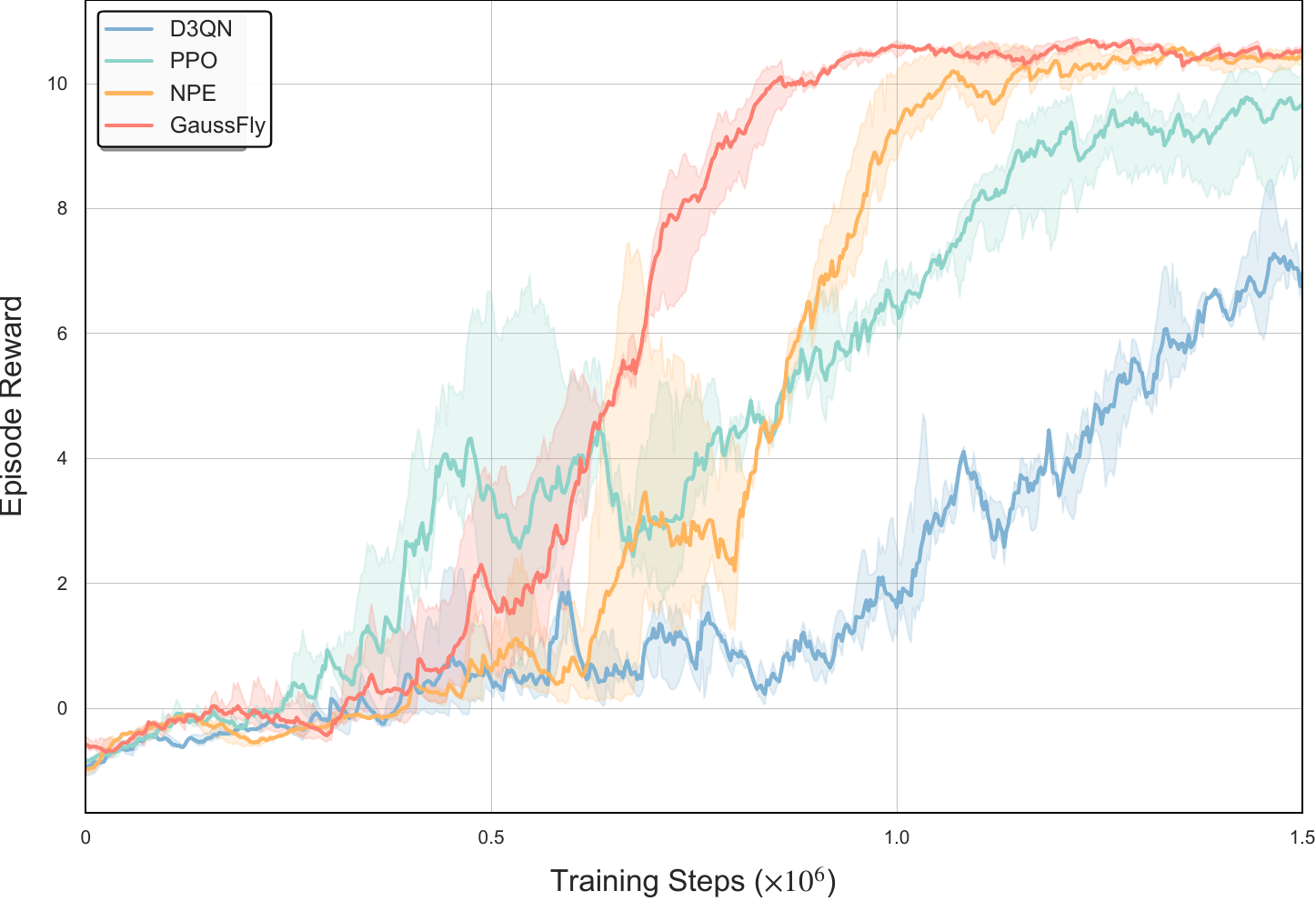}
    \caption{Training reward curves. GaussFly achieves comparable asymptotic performance to NPE but exhibits significantly higher sample efficiency. Although NPE benefits from prior guidance, its direct reliance on raw RGB inputs still bottlenecks its learning speed. Furthermore, the standard PPO and D3QN baselines show inferior sample efficiency and asymptotic returns due to the lack of specialized representation mechanisms in vanilla RL algorithms.}
    \label{fig:reward}
\end{figure}

\begin{table}[t!]
\centering
\caption{Performance of Baselines in the Unseen Evaluation Environment}
\label{tab:baseline}
\setlength{\tabcolsep}{3pt} 
\resizebox{\columnwidth}{!}{%
\renewcommand{\arraystretch}{1.15} 
\begin{tabular}{lcccccc}
\toprule
Method & NE$\downarrow$ & OS$\uparrow$ & SR$\uparrow$ & SPL$\uparrow$ & CR$\downarrow$ & TTS$\downarrow$ \\
\midrule
D3QN~\cite{zhang2023partially}    & 2.06$\pm$0.55 & 46.0$\pm$6.0 & 42.5$\pm$5.0  & 0.34$\pm$0.06 & 53.0$\pm$5.5 & 398$\pm$13 \\
PPO~\cite{schulman2017proximal}   & 1.72$\pm$0.47 & 57.5$\pm$5.0 & 51.5$\pm$4.0  & 0.36$\pm$0.06 & 40.5$\pm$4.5 & 373$\pm$11 \\
NPE~\cite{zhang2024npe}           & 1.04$\pm$0.31 & 64.0$\pm$4.0 & 60.5$\pm$3.5  & 0.48$\pm$0.04 & 36.0$\pm$4.0 & 329$\pm$9 \\
DAgger                            & 1.27$\pm$0.33 & 60.5$\pm$4.5 & 56.5$\pm$4.0  & 0.46$\pm$0.04 & 42.0$\pm$4.5 & 341$\pm$11 \\
Hybrid-APF~\cite{pan2021improved} & 0.84$\pm$0.54 & 74.5$\pm$6.5 & 70.0$\pm$6.0  & 0.68$\pm$0.06 & 12.0$\pm$10.0 & 353$\pm$17 \\
GaussFly (Ours)                              & \textbf{0.52$\pm$0.24} & \textbf{88.5$\pm$3.0} & \textbf{84.0$\pm$2.0} & \textbf{0.81$\pm$0.03} & \textbf{9.5$\pm$2.0} & \textbf{306$\pm$7} \\
\bottomrule
\end{tabular}%
}
\end{table}

\section{Results and Analysis}
This section presents the experimental results of GaussFly, including simulation results, real-world flight evaluations, and ablation studies on the key components.

\subsection{Simulation Results}

We conduct training in Scene A and perform zero-shot evaluation in Scene B without fine-tuning to systematically evaluate the effectiveness of GaussFly. The training reward curves are shown in Fig. \ref{fig:reward}. Note that Hybrid-APF and DAgger are excluded from the reward curve comparisons; Hybrid-APF operates as a planning-based baseline, while DAgger trains via expert-guided IL rather than reward maximization.

\begin{figure}[t!]
    \centering
    \includegraphics[width=1.0\linewidth]{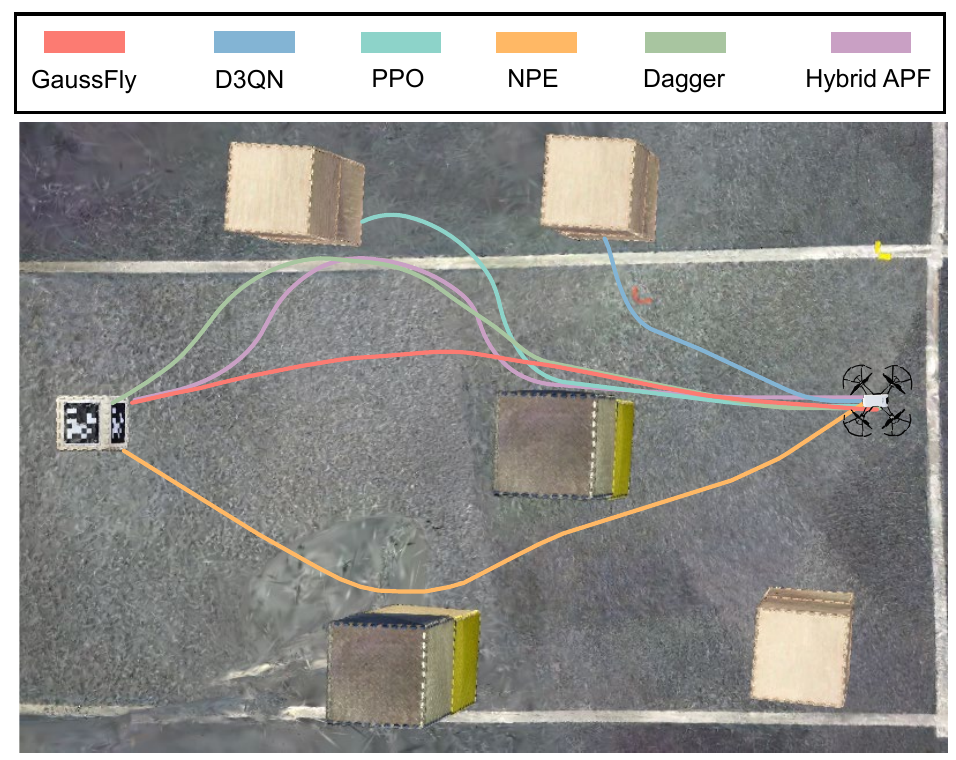}
    \caption{Visualization of representative flight trajectories. GaussFly (red) generates the smoothest and most direct path in the unseen environment, whereas baselines suffer from collisions or highly curved detours.}
    \label{fig:tra_sim}
\end{figure}

\begin{figure*}[t!]
   \centering
   \includegraphics[width=1.0\textwidth]{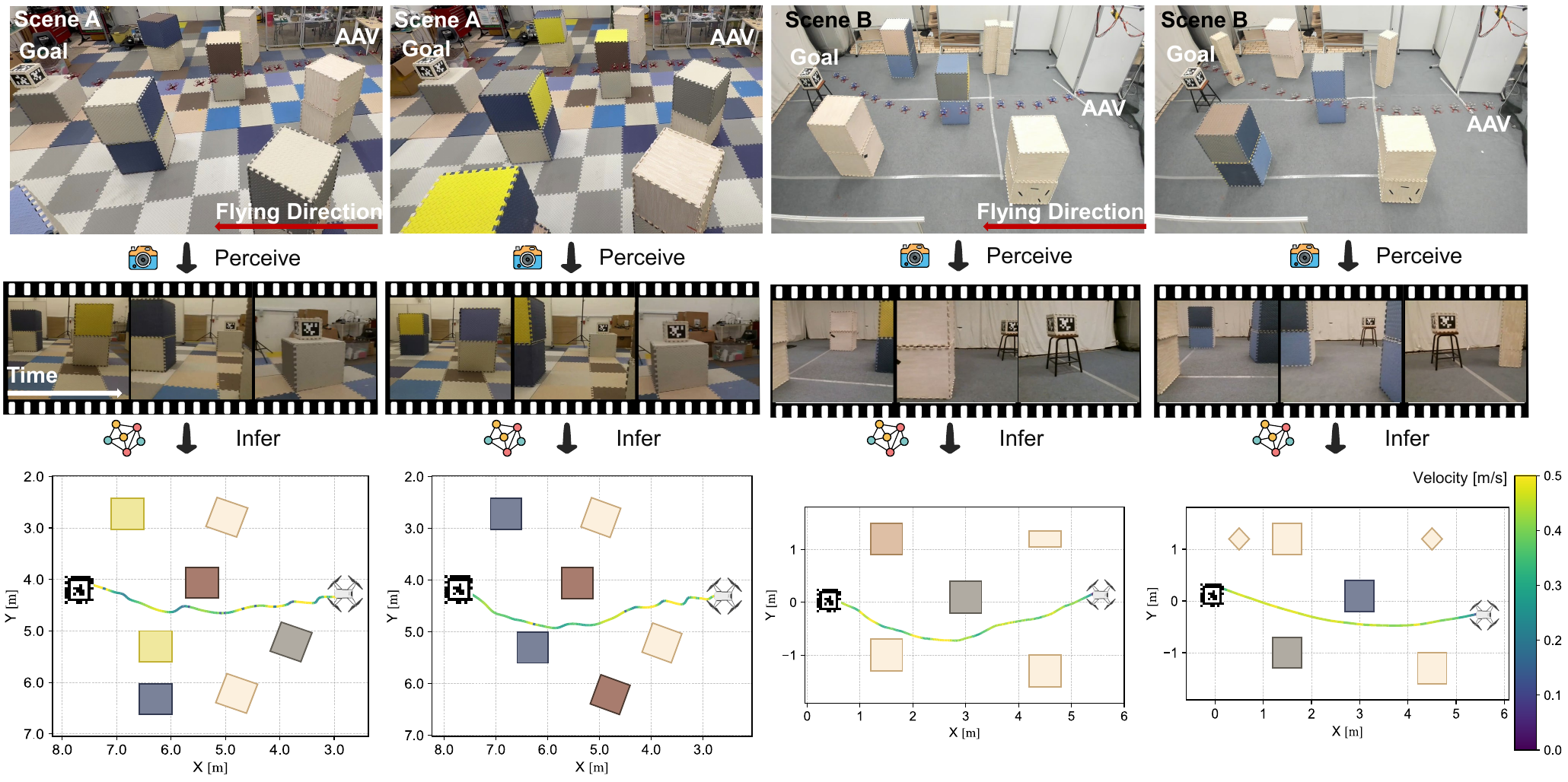} 
   \caption{Representative real-world flight trajectories in two indoor environments. Top: Third-person views of the AAV navigating towards the goal. Middle: Consecutive monocular FPV observations captured onboard. Bottom: Executed 2D flight trajectories with velocity profiles. Despite varying obstacle layouts and visual textures, GaussFly achieves robust and collision-free zero-shot navigation across environments.}
   \label{fig:tra_physical}
\end{figure*}

\textbf{(1) Training Results.} For training, our method achieves a comparable asymptotic performance to NPE, while exhibiting substantially higher sample efficiency. This improvement can be attributed to the use of a pre-trained visual encoder obtained through contrastive representation learning. Instead of directly operating on high-dimensional RGB observations, our policy network receives compact and low-dimensional latent features, which effectively reduce computational overhead and facilitate more sample-efficient policy learning. Although NPE accelerates early learning by leveraging non-expert demonstrations, it still relies on consecutive raw RGB inputs. As a result, it suffers from increased sample complexity and slower convergence. Furthermore, the standard PPO and D3QN baselines exhibit inferior performance in both sample efficiency and asymptotic performance. As vanilla RL algorithms lack specialized representation mechanisms, they struggle to effectively map high-dimensional visual observations to control commands, leading to suboptimal policy learning.

\textbf{(2) Evaluation Results.}To further evaluate zero-shot generalization, we test all baselines in Scene B without any fine-tuning. Quantitative results are summarized in Table \ref{tab:baseline}. GaussFly consistently outperforms all baselines across all metrics, demonstrating superior cross-environment navigation robustness. Specifically, while NPE exhibits a certain degree of generalization due to its transition to independent exploration in later training stages, its reliance on raw RGB inputs still bottlenecks its adaptability to unseen scenes. Similarly, as vanilla RL methods lack invariant feature extraction, D3QN and PPO severely overfit the training environment; their performance degrades significantly when deployed in the scene with unseen layouts and textures. DAgger also exhibits a notable performance drop, as imitation-based policies inherently overfit to prior demonstrations and struggle with domain shifts. In contrast, Hybrid-APF achieves comparable performance, reflecting the inherent robustness of planning-based approaches. Nevertheless, its overall performance remains inferior to GaussFly, as potential field-based planners are prone to local minima, whereas our contrastive-aligned visual representations provide richer and more resilient environmental awareness. Representative flight trajectories are visualized in Fig. \ref{fig:tra_sim}. GaussFly produces the smoothest and most direct trajectories in the evaluation environment. In comparison, other baselines either collide with obstacles or generate highly curved trajectories in the unseen environment, highlighting the advantage of extracting robust latent representations from RGB observations through contrastive learning.

\subsection{Real-World Experiments}

We conduct physical flight experiments in Scenes A and B to further assess the sim-to-real gap and the real-world transferability of GaussFly. Notably, we directly deploy the learned policy into the physical world without fine-tuning, even though part of the obstacle configurations and textures differ from those used during simulation training. These visual variations inherently increase the sim-to-real gap, providing a rigorous evaluation of the learned policy's adaptability.

The AAV platform used for evaluation is a DJI Tello Edu quadrotor equipped with a forward-facing monocular RGB camera streaming $720p$ video at $30$ FPS with a field of view of $82.6^\circ$. External localization is provided by an OptiTrack motion capture system and a UWB positioning system. Before each trial, start and goal positions are manually specified. During flight, onboard RGB observations are resized and processed through the pretrained visual encoder to extract compact features. These features, together with the proprioceptive state obtained from the localization system, are fed into the policy network to generate continuous control commands. The complete model is exported in ONNX format and deployed for real-time inference without additional fine-tuning.

\begin{figure}[htbp]
    \centering
    \includegraphics[width=0.9\linewidth]{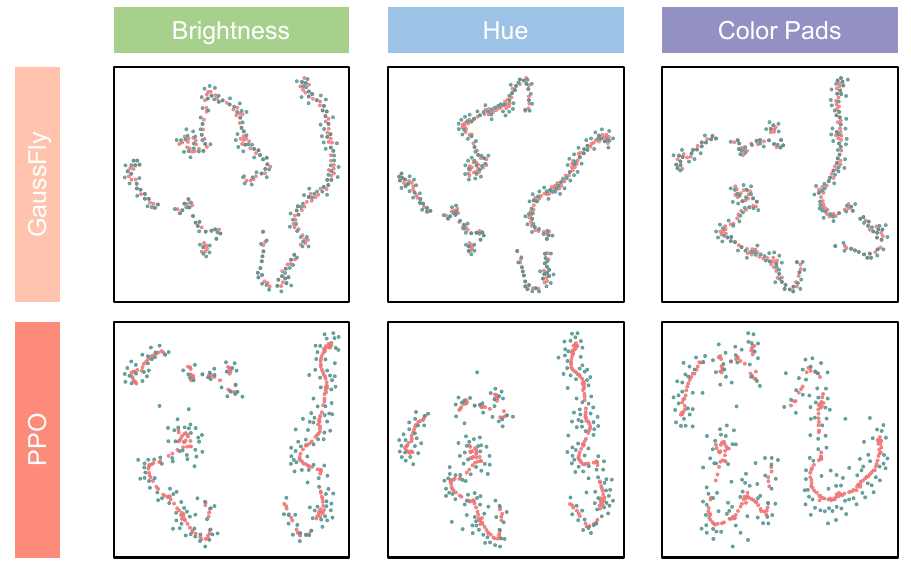}
    \caption{t-SNE visualization under visual interference. Red and green dots denote features from the original and perturbed images, respectively. Our contrastive-based encoder yields tightly aligned feature clusters regardless of visual noise, indicating highly invariant representations. Conversely, the features generated by PPO are highly scattered and disorganized, demonstrating a strong vulnerability to domain shifts.}
    \label{fig:tsne}
\end{figure}

\begin{figure}[t!]
    \centering
    \includegraphics[width=0.9\linewidth]{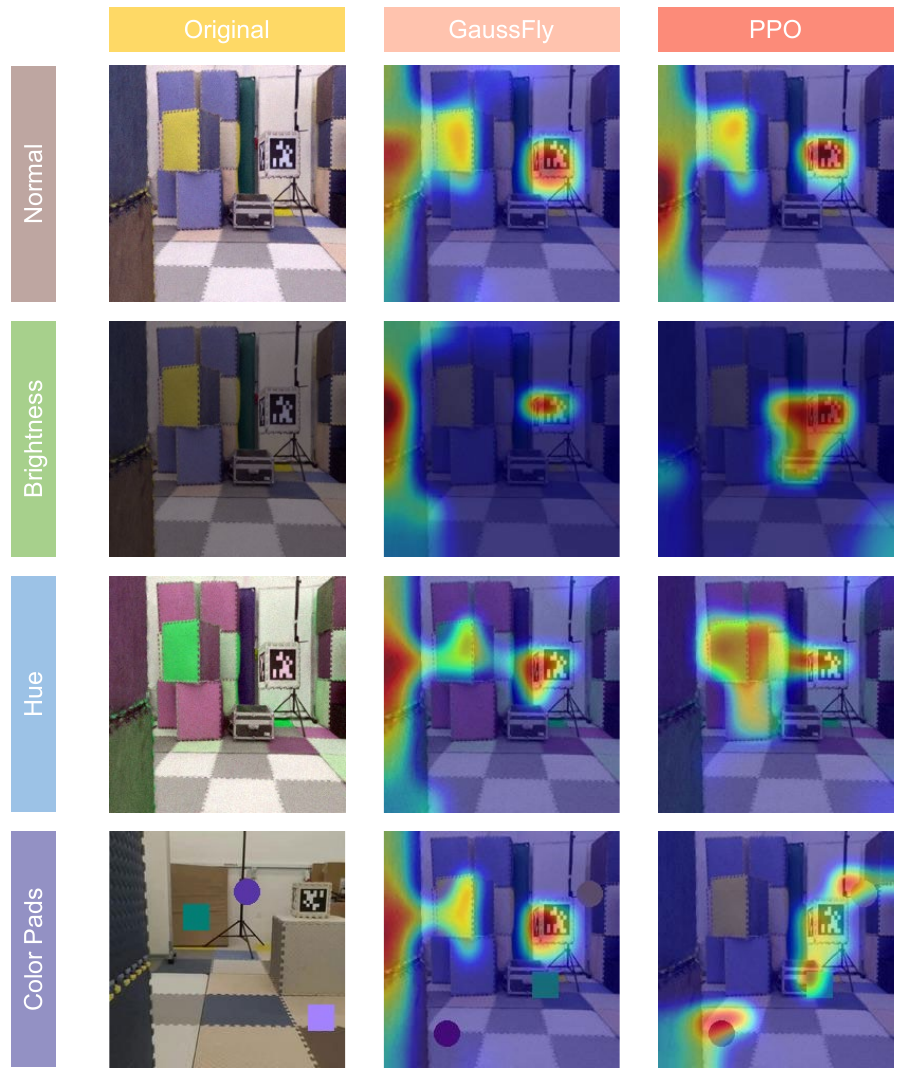}
    \caption{Attention visualization in simulation. Our contrastive-based encoder consistently maintains a precise focus on essential geometric features, such as obstacles and targets, regardless of brightness, hue, or artificial color pad perturbations. Conversely, PPO demonstrates high vulnerability to domain shifts, often misdirecting its attention toward irrelevant textures or scattering across the background when image properties are altered.}
    \label{fig:att_sim}
\end{figure}

\begin{figure}[t!]
    \centering
    \includegraphics[width=0.9\linewidth]{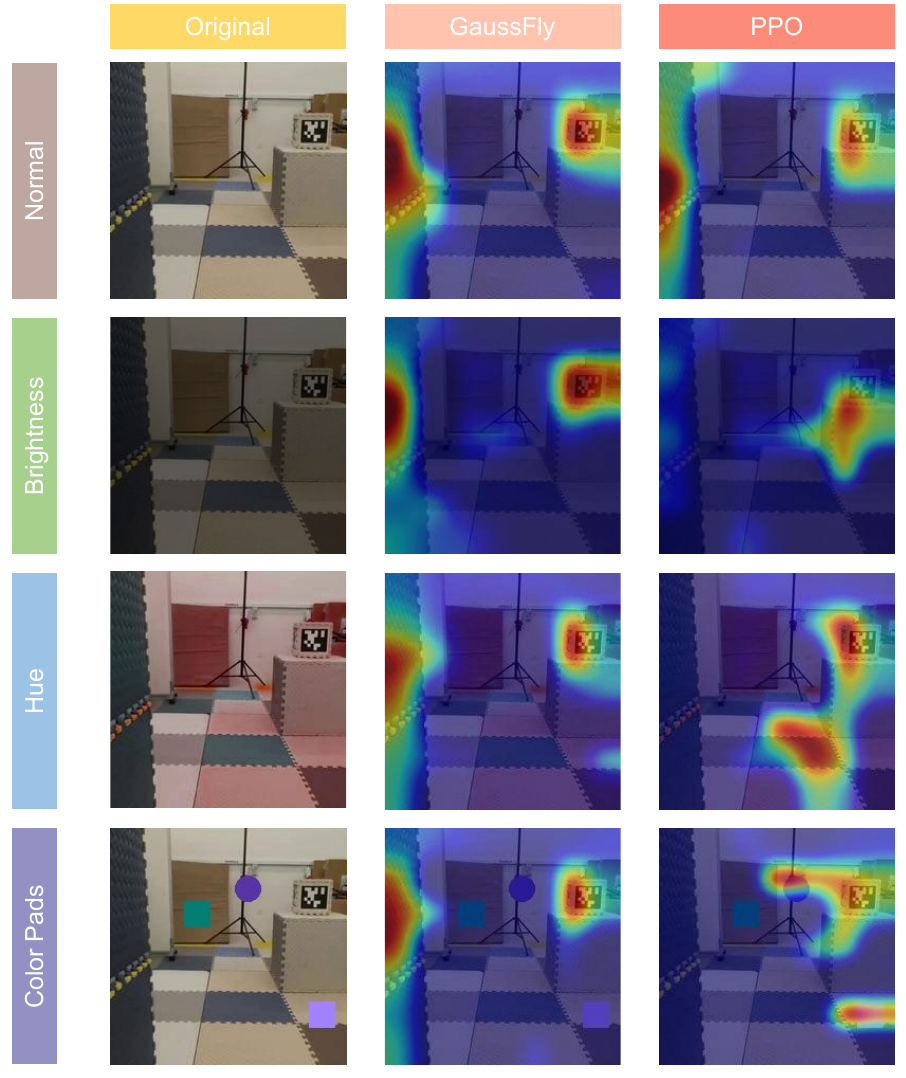}
    \caption{Attention visualization in real-world environments. By combining 3DGS-based training environments with contrastive representation learning, GaussFly robustly anchors its attention on task-critical physical objects. Conversely, PPO is more easily distracted by visual interference as it lacks this noise-resilient feature extraction.}
    \label{fig:att_real}
\end{figure}

We evaluate GaussFly over 20 trials per environment under varying start--goal configurations. In Scene A, which corresponds to the simulation training environment, GaussFly achieves a success rate of 80\%. In the unseen Scene B, the success rate is 65\%. Representative flight trajectories are shown in Fig.~\ref{fig:tra_physical}. Despite changes in obstacle texture and layout, GaussFly maintains stable and goal-directed behavior. These results indicate that the sim-to-real gap is effectively reduced by two key components of GaussFly. (i) The contrastive learning objective ensures that the learned representations capture task-relevant information rather than relying on environment-specific visual textures. (ii) The reconstructed 3DGS environment provides photorealistic and geometrically accurate training observations, which inherently minimizes the visual gap between simulation and the real world.

\subsection{Representation Analysis}
We conduct an analysis of the learned representations through t-SNE embedding \cite{van2008visualizing} and attention visualization, providing insights into the superior performance of GaussFly.

We utilize t-SNE to visualize the distribution of latent features and evaluate the robustness of the learned representations against visual interference. Specifically, we introduce random perturbations to the rendered images: brightness is scaled within $[0.5, 1.5]$, hue is shifted within $[-0.5, 0.5]$, and random color patches are overlaid on the background. We then compare the feature clusters of the original and perturbed images for GaussFly and the standard PPO baseline. As illustrated in Fig. \ref{fig:tsne}, GaussFly maintains tight clustering between the original and perturbed samples, indicating that the contrastive encoder has successfully learned invariant features resilient to visual noise. In contrast, the latent features extracted by the PPO baseline scatter significantly under perturbation, revealing its inherent sensitivity to visual variations and the fragility of its end-to-end representations.

To further investigate the interpretability of the learned representations, we employ Grad-CAM \cite{selvaraju2017grad} to visualize the regions of interest contributing to the agent's decision-making. We generate activation heatmaps for observations from both the simulation and real-world environments, where warmer colors indicate higher importance. As shown in Figs. \ref{fig:att_sim} and \ref{fig:att_real}, our feature encoder consistently focuses on task-relevant geometric structures, specifically obstacles and the target region. Crucially, this attention remains highly stable even in the presence of complex backgrounds or simulated visual perturbations. Conversely, the standard PPO baseline fails to exhibit such selective focus. Its attention is frequently distracted by irrelevant background textures or dispersed irregularly across the image, lacking the spatial awareness required for safe collision avoidance. This significant contrast explains GaussFly’s superior ability to generalize to unseen environments.

\begin{figure}[t!]
    \centering
    \includegraphics[width=0.9\linewidth]{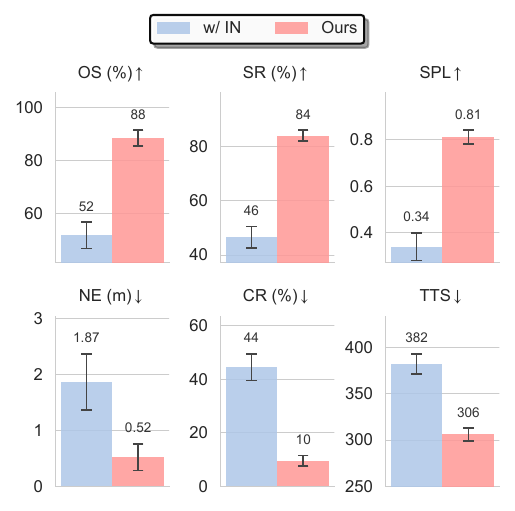}
    \caption{Ablation study on contrastive pre-training. We evaluate navigation performance by comparing our contrastively learned encoder against a baseline using frozen ImageNet weights (w/ IN). Our framework achieves superior overall performance. In contrast, the w/ IN variant leads to significant performance degradation, as standard classification pre-training fails to distill the task-relevant and noise-robust features required for complex visuomotor policies.}
    \label{fig:abl_contra}
\end{figure}

\begin{figure}[t!]
    \centering
    \includegraphics[width=0.9\linewidth]{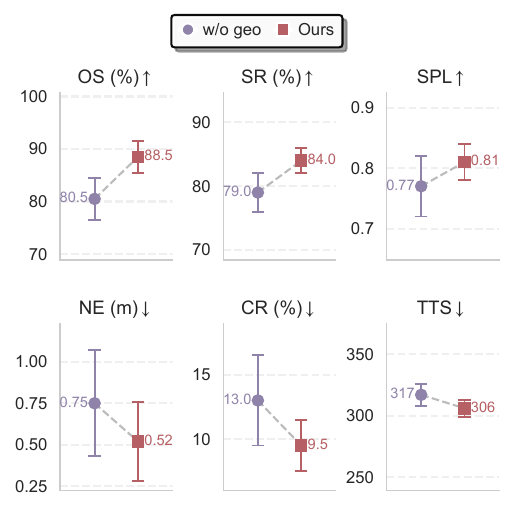}
    \caption{Ablation study on geometric constraints. We evaluate navigation performance by comparing our geometrically consistent 3DGS framework against a vanilla 3DGS baseline (w/o geo). Our framework achieves superior overall performance. In contrast, the w/o geo variant leads to increased collision rates and general performance degradation, as unconstrained 3DGS generates structural artifacts and depth inconsistencies that impair precise perception.}
    \label{fig:abl_3dgs}
\end{figure}

\subsection{Ablation Studies}
To further validate the contribution of our core framework designs, we conduct ablation studies on the feature extractor and the 3DGS simulation environment. Here, we only report the evaluation results, as the navigation performance of the visuomotor policy is our primary focus.

\textbf{(1) Impact of Contrastive Pre-training.} We investigate the impact of our contrastive representation learning on navigation performance. Specifically, we train a visuomotor policy using a standard ResNet-50 initialized with ImageNet weights as the feature extractor, with its weights frozen during policy training (denoted as w/ IN). The resulting policy is then evaluated in Scene B. As shown in Fig. \ref{fig:abl_contra}, the w/ IN variant leads to a significant performance drop. Because ImageNet features are inherently optimized for general computer vision tasks, such as object classification, they are not further optimized to extract the task-relevant and robust representations required for visuomotor policies. This impairs the model's ability to reliably process complex visual observations during flight. These results confirm the utility of our contrastive learning framework. Our pre-training objective encourages the encoder to distill task-relevant features that are robust to visual noise, whereas standard classification pre-training does not explicitly optimize for such invariance.



\textbf{(2) Impact of Geometric Constraints.} We investigate the impact of applying geometric constraints during 3DGS reconstruction on navigation performance. Specifically, we train a visuomotor policy in a vanilla 3DGS environment optimized without the geometric regularization term, denoted as w/o geo. The resulting policy is then evaluated in Scene B. As shown in Fig. \ref{fig:abl_3dgs}, the w/o geo variant leads to a performance drop and an increased collision rate. We attribute this degradation to the fact that the original 3DGS prioritizes visual fidelity over structural accuracy, frequently generating irregular surface artifacts and depth inconsistencies. This mismatch between visual rendering and actual physical boundaries impairs the model's ability to learn precise collision avoidance from visual inputs. These results confirm the utility of our geometrically consistent 3DGS framework. Explicitly enforcing geometric constraints ensures that the reconstructed simulation provides structurally accurate visual observations, while unconstrained 3DGS degrades navigation performance due to misleading spatial representations.



\section{Conclusion}
In this work, we propose GaussFly, a novel visuomotor policy learning framework that explicitly decouples representation learning from policy learning. By combining geometrically consistent 3DGS reconstructions with contrastive representation learning, GaussFly enables robust, monocular autonomous navigation for AAVs. The visual encoder is contrastively pre-trained to extract compact, noise-resilient latent features. This encoder is subsequently frozen and used to provide low-dimensional feature inputs for visuomotor policy learning, which significantly improves sample efficiency and asymptotic performance. Extensive simulation and real-world experiments demonstrate that GaussFly effectively bridges the sim-to-real gap, enabling robust zero-shot policy transfer across unseen physical scenes with complex layouts and textures.

However, a notable limitation is the static illumination inherent to standard 3DGS. Because lighting is baked into the radiance fields, illumination variations can only be approximated via 2D image-level perturbations. To address this residual domain gap, our future work will focus on integrating evolvable simulation environments with physically-based relighting capabilities. Additionally, we plan to extend GaussFly to vision-language navigation tasks, where grounding visual observations to language instructions remains a challenge.

\bibliographystyle{IEEEtran}
\bibliography{IEEEabrv,yuhang}

\vspace{-1cm}
\begin{IEEEbiography}[{\includegraphics[width=1in,height=1.25in,clip,keepaspectratio]{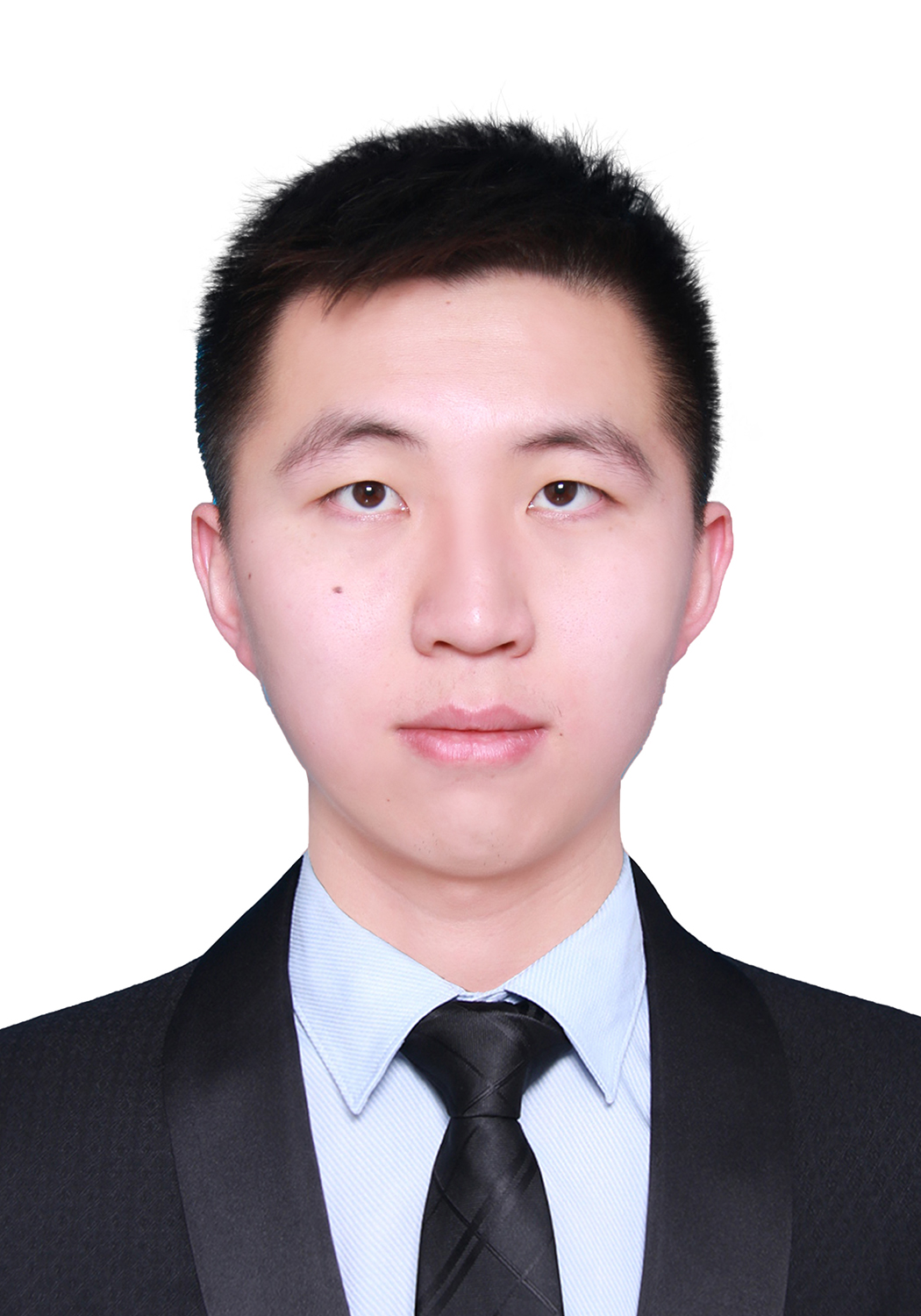}}]{Yuhang Zhang} (graduate student member, IEEE) received the B.E. degree in flight vehicle propulsion engineering from Harbin Engineering University, Harbin, China, in 2021 and the M.Eng. degree from the School of Mechanical \& Aerospace Engineering at Nanyang Technological University (NTU), Singapore, in 2023. Currently,  he is pursuing his Ph.D. degree at the School of Mechanical \& Aerospace Engineering at NTU, Singapore.

His research primarily focuses on unmanned aerial vehicles, deep reinforcement learning, and vision-and-language navigation.
\end{IEEEbiography}

\begin{IEEEbiography}[{\includegraphics[width=1in,height=1.25in,clip,keepaspectratio]{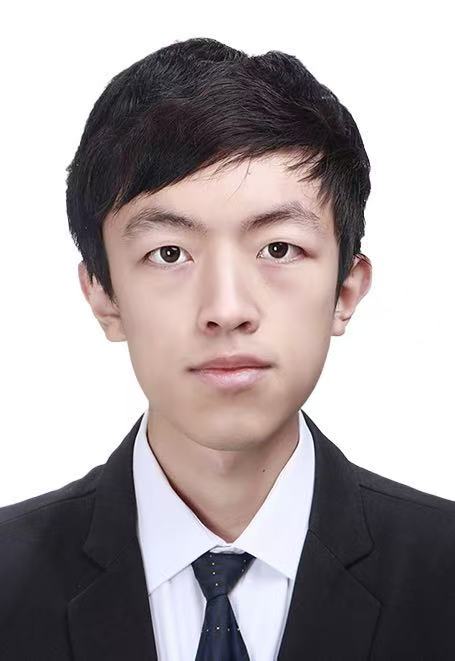}}]{Mingsheng Li} (graduate student member, IEEE) received the B.E. degree in aerospace engineering from Beihang University, Beijing, China, in 2022, and M.S. degree from Imperial College London, London, U.K., in 2023. He is currently working toward the Ph.D. degree in mechanical and aerospace engineering with Nanyang Technological University (NTU), Singapore.

His research interests include multi-agent systems, unmanned aerial vehicles and reinforcement learning.
\end{IEEEbiography}

\begin{IEEEbiography}[{\includegraphics[width=1in,height=1.25in,clip,keepaspectratio]{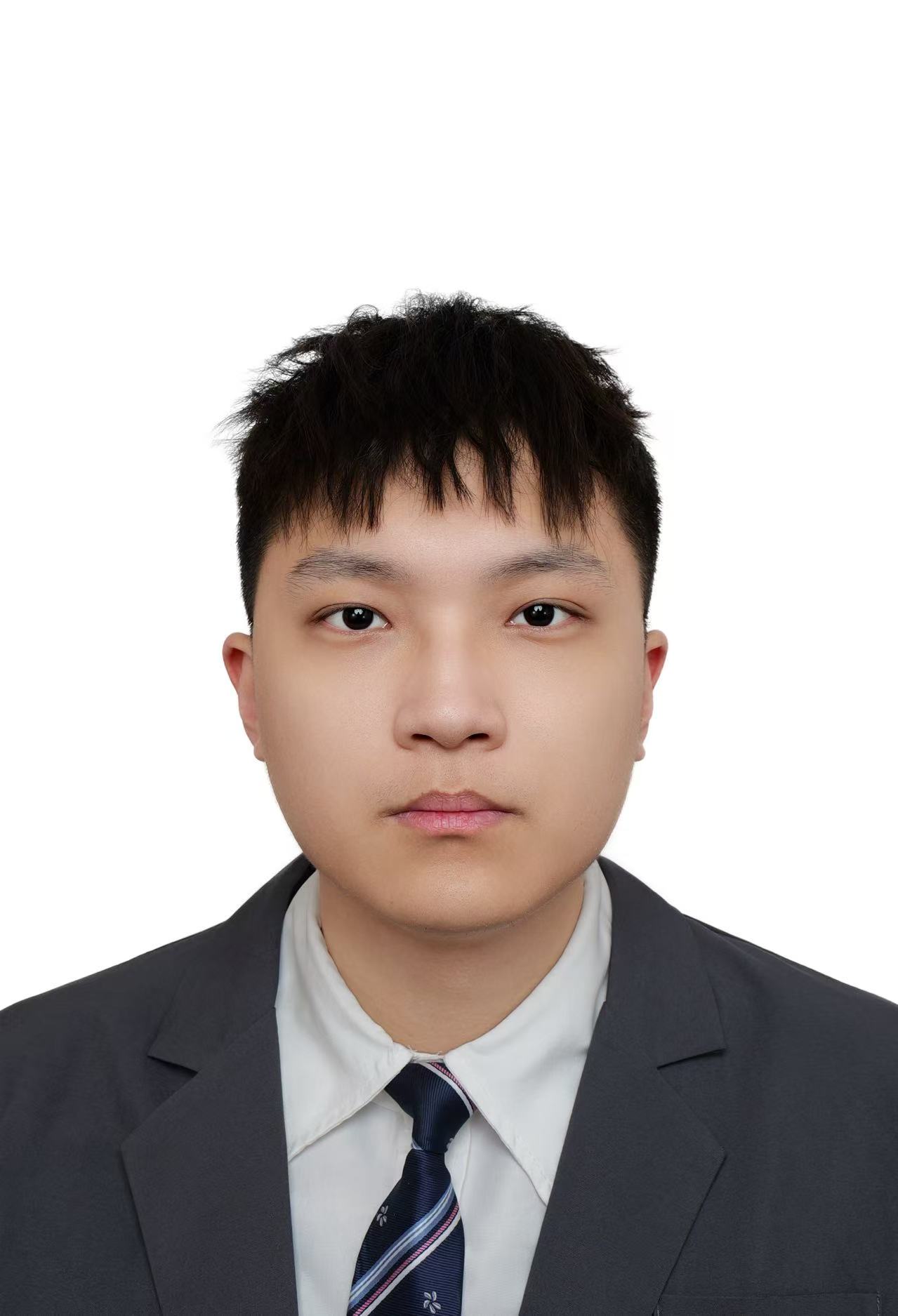}}]{Yujing Shang} received the B.E. degree in electronic information science and technology from the Hefei University of Technology, Hefei, China, in 2024. He is currently pursuing the M.Sc. degree in computer control and automation with the School of Electrical and Electronic Engineering, Nanyang Technological University (NTU), Singapore.

His research interests include unmanned aerial vehicles, multi-agent reinforcement learning, and 3D Gaussian splatting for simulation environments.
\end{IEEEbiography}

\begin{IEEEbiography}[{\includegraphics[width=1in,height=1.25in,clip,keepaspectratio]{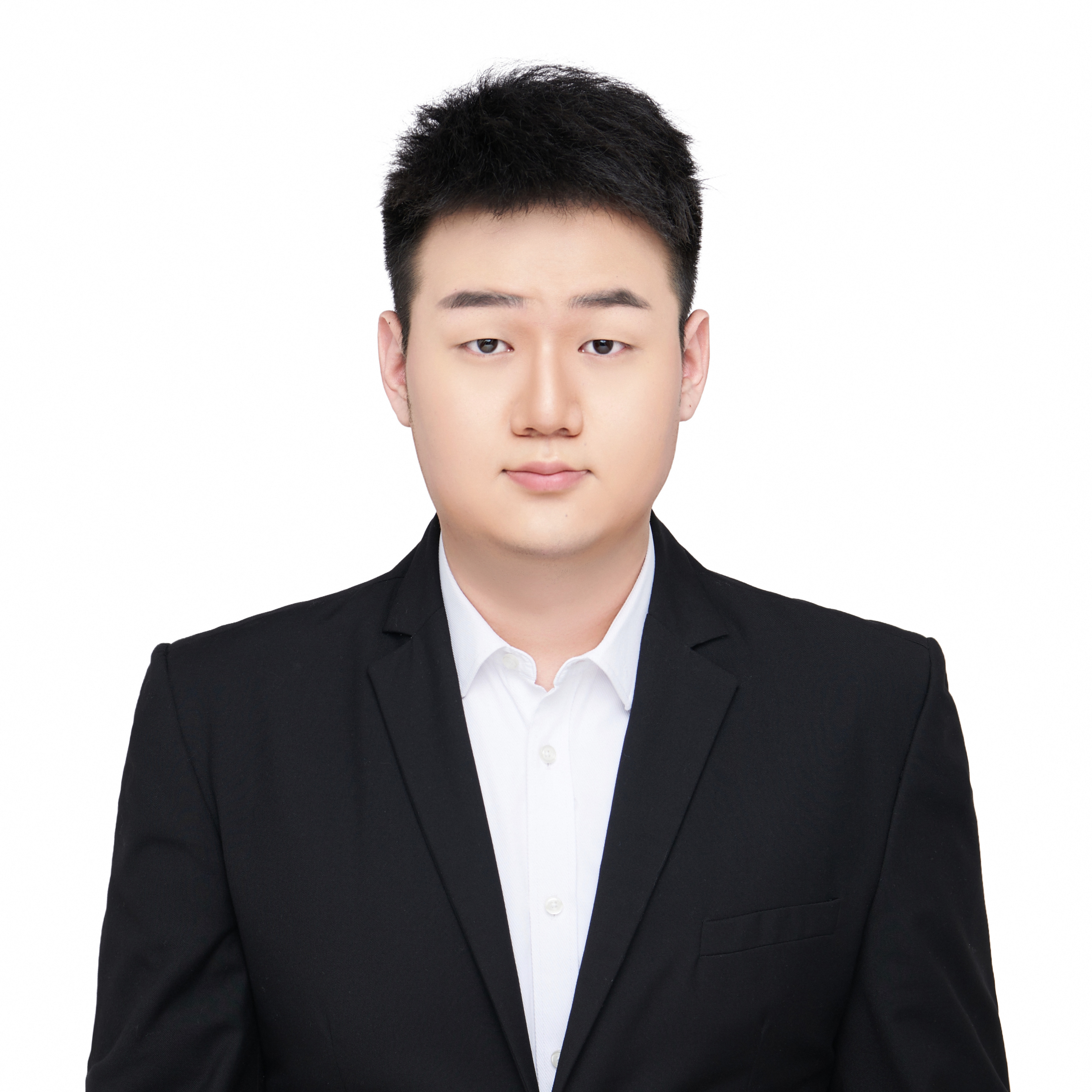}}]{Zhuoyuan Yu} (graduate student member, IEEE) received the B.E. degree from School of Aeronautics at Northwestern Polytechnical University, Xi'an, China, in 2023 and the M.Eng. degree from the College of Design and Engineering at National University of Singapore, Singapore, in 2025. Currently, he is a research intern in Dexmal, Beijing, China.

His research interests include deep reinforcement learning, vision-language-action models, and navigation.
\end{IEEEbiography}

\begin{IEEEbiography}[{\includegraphics[width=1in,height=1.25in,clip,keepaspectratio]{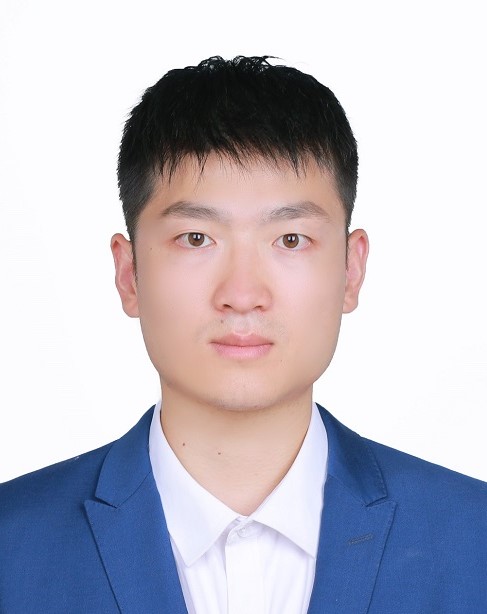}}]{Chao Yan} (member, IEEE) received the B.E. degree in electrical engineering and automation from China University of Mining and Technology, Xuzhou, China, in 2017, and the M.S. and Ph.D. degrees in control science and engineering from the National University of Defense Technology, Changsha, China, in 2019, and 2023, respectively. He was a visiting Ph.D. student with the School of Mechanical and Aerospace Engineering, Nanyang Technological University, Singapore, from 2021 to 2022.

He is currently an Associate Professor with the College of Automation Engineering, Nanjing University of Aeronautics and Astronautics, Nanjing, China. His research interests include deep reinforcement learning and coordination control of UAV swarms.
\end{IEEEbiography}

\begin{IEEEbiography}[{\includegraphics[width=1in,height=1.25in,clip,keepaspectratio]{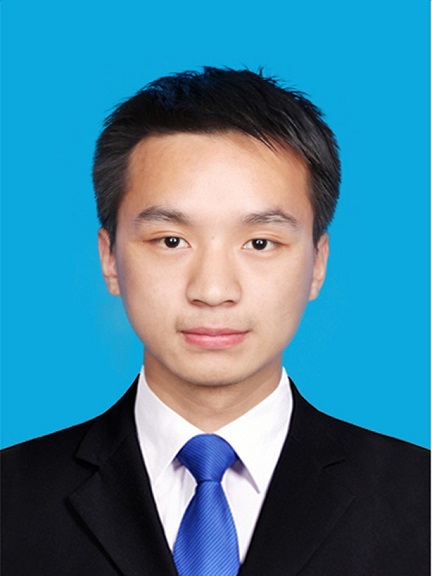}}]{Jiaping Xiao} (member, IEEE)
received the B.E. degree in aircraft design and engineering and the M.S. degree in flight dynamics and control from Beihang University, Beijing, China, in 2014 and 2017, and the Ph.D. degree in intelligent systems from Nanyang Technological University (NTU), Singapore, in 2024. He is currently a research fellow with the School of Mechanical and Aerospace Engineering, NTU.

His research interests include cyber-physical systems, reinforcement learning, machine vision, and aerial robotics.
\end{IEEEbiography}

\begin{IEEEbiography}[{\includegraphics[width=1in,height=1.25in,clip,keepaspectratio]{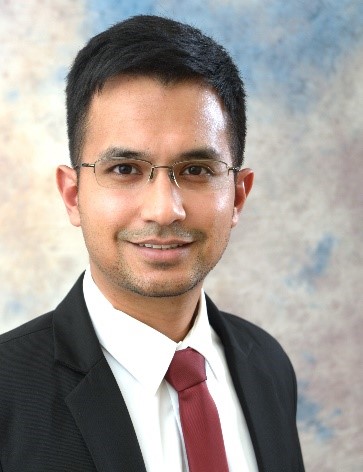}}]{Mir Feroskhan}
(member, IEEE) received B.E. degree (Hons.) in aerospace engineering from Nanyang Technological University, Singapore, in 2011, and the Ph.D. degree in aerospace engineering from the Florida Institute of Technology, Melbourne, FL, in 2016. He is currently an assistant professor with the School of Mechanical \& Aerospace Engineering at NTU.

His research interests include nonlinear control systems, multi-agent systems, flight dynamics and control, and aerial robotics.
\end{IEEEbiography}

\end{document}